\documentclass[conference]{IEEEtran}
\usepackage{todonotes}

\usepackage{tabularray}
\usepackage{graphicx}
\usepackage{float}
\usepackage{amsmath,amssymb,amsfonts}
\usepackage{algorithmic}
\usepackage{textcomp}
\usepackage{xcolor}
\usepackage{adjustbox}
\usepackage{float}
\usepackage[utf8]{inputenc}
\usepackage{caption}
\usepackage{comment}
\usepackage{wrapfig}
\usepackage[compact]{titlesec}
\usepackage{flushend}
\usepackage[colorlinks=true,urlcolor=black]{hyperref}
\AtBeginDocument{%
  \providecommand\BibTeX{{%
    \normalfont B\kern-0.5em{\scshape i\kern-0.25em b}\kern-0.8em\TeX}}}

\IEEEoverridecommandlockouts
\begin{document}

\title{Exploring the Performance of ML/DL Architectures on the MNIST-1D Dataset}

\makeatletter 
\newcommand{\linebreakand}{%
  \end{@IEEEauthorhalign}
  \hfill\mbox{}\par
  \mbox{}\hfill\begin{@IEEEauthorhalign}
}
\makeatother 

\author{

\IEEEauthorblockN{Michael Beebe}
\IEEEauthorblockA{\textit{Department of Computer Science} \\
\textit{Texas Tech University}\\
Lubbock, TX \\
michael.beebe@ttu.edu}

\and
\IEEEauthorblockN{GodsGift Uzor}
\IEEEauthorblockA{\textit{Department of Computer Science} \\
\textit{Texas Tech University}\\
Lubbock, TX \\
godsgift.uzor@ttu.edu}

\linebreakand

\and
\IEEEauthorblockN{Manasa Chepuri}
\IEEEauthorblockA{\textit{Department of Computer Science} \\
\textit{Texas Tech University}\\
Lubbock, TX  \\
mchepuri@ttu.edu}


\and
\IEEEauthorblockN{Divya Sree Vemula}
\IEEEauthorblockA{\textit{Department of Computer Science} \\
\textit{Texas Tech University}\\
Lubbock, TX  \\
divemula@ttu.edu}

\and
\IEEEauthorblockN{Angel Ayala}
\IEEEauthorblockA{\textit{Department of Computer Science} \\
\textit{Texas Tech University}\\
Lubbock, TX  \\
angel.ayala@ttu.edu}

}



\maketitle

\begin{abstract}
Small datasets like MNIST have historically been instrumental in advancing machine learning research by providing a controlled environment for rapid experimentation and model evaluation. However, their simplicity often limits their utility for distinguishing between advanced neural network architectures. To address these challenges, Greydanus et al. introduced the MNIST-1D dataset, a one-dimensional adaptation of MNIST designed to explore inductive biases in sequential data. This dataset maintains the advantages of small-scale datasets while introducing variability and complexity that make it ideal for studying advanced architectures.

In this paper, we extend the exploration of MNIST-1D by evaluating the performance of Residual Networks (ResNet), Temporal Convolutional Networks (TCN), and Dilated Convolutional Neural Networks (DCNN). These models, known for their ability to capture sequential patterns and hierarchical features, were implemented and benchmarked alongside previously tested architectures such as logistic regression, MLPs, CNNs, and GRUs. Our experimental results demonstrate that advanced architectures like TCN and DCNN consistently outperform simpler models, achieving near-human performance on MNIST-1D. ResNet also shows significant improvements, highlighting the importance of leveraging inductive biases and hierarchical feature extraction in small structured datasets.

Through this study, we validate the utility of MNIST-1D as a robust benchmark for evaluating machine learning architectures under computational constraints. Our findings emphasize the role of architectural innovations in improving model performance and offer insights into optimizing deep learning models for resource-limited environments.
\end{abstract}



\section{Introduction}
In the field of machine learning, small datasets have proven to be crucial for exploring foundational concepts with minimal computational overhead. These datasets allow researchers to prototype models, tune hyperparameters, and experiment with new ideas in a controlled environment. The MNIST dataset \ref{fig:raw-data} \cite{greydanus2020scaling}, for example, has become a standard benchmark in the development of deep learning architectures, owing to its small size and ease of use. Training on MNIST requires only modest computational resources, making it accessible for both educational purposes and cutting-edge research.

Small datasets like MNIST offer significant advantages. Their structured format and small size enable rapid experimentation, which is critical for developing and testing new models. Researchers can run many training steps in a short period of time, facilitating faster discovery and refinement of ideas. Additionally, small datasets require less memory and computational resources, allowing for experimentation on a wider range of hardware, from high-performance servers to personal computers, making machine learning research more accessible to a broader audience.

\begin{figure*}[!htb]
    \centering    \includegraphics[width=.80\linewidth]{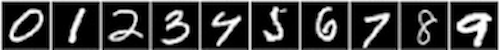}
    \caption{Original MNIST digit images in 2D grayscale format \cite{original_mnist}}
    \label{fig:raw-data}
\end{figure*}

\begin{figure*}[!htb]
    \centering
    \includegraphics[width=.80\linewidth]{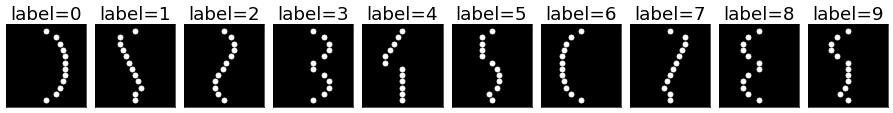}
    \caption{Sparse 1D digit templates using 12 x-coordinates for each digit (0–9) \cite{greydanus2020scaling}}
    \label{fig:black-signals}
\end{figure*}

\begin{figure*}[!htb]
    \centering
    \includegraphics[width=.80\linewidth]{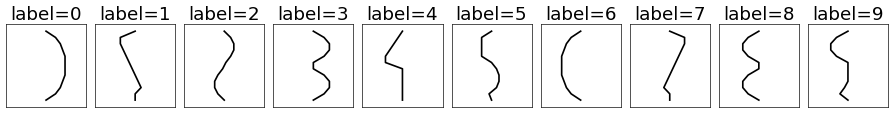}
    \caption{Digit Templates after padding, adding noise, scaling, and Gaussian smoothing \cite{greydanus2020scaling}}
    \label{fig:white-signals}
\end{figure*}

\begin{figure*}[!htb]
    \centering
    \includegraphics[width=.80\linewidth]{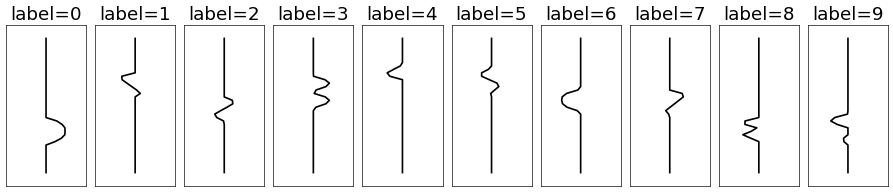}
    \caption{Final downsampled MNIST-1D Digit Representations with 40 points \cite{greydanus2020scaling}}
    \label{fig:transformed-signals}
\end{figure*}

Furthermore, datasets like MNIST serve as excellent benchmarks for comparing different models. They provide a common ground where researchers can test models and evaluate performance in a consistent, reproducible manner. Because of their simplicity, datasets like MNIST allow for focused investigation into specific model characteristics, such as how well a model exploits inductive biases or responds to various regularization techniques. These insights can then be applied when scaling up to larger, more complex datasets.

However, small datasets like MNIST also present challenges. One of the main reasons for this is the limited complexity and variability in the data. The MNIST dataset, for instance, consists of simple, grayscale images with clear and distinguishable features, allowing even basic machine-learning models to achieve high accuracy. As a result, many models, whether simple or complex, tend to perform similarly well on MNIST. Since the dataset’s structure is relatively straightforward, models do not need to exploit advanced inductive biases, which can obscure the benefits of more sophisticated architectures such as deep CNNs or regularization techniques designed for more challenging datasets. Furthermore, this lack of complexity makes it difficult to differentiate between models’ performance and fully appreciate the strengths of advanced neural networks.

\begin{figure}[!htb]
    \centering
    \includegraphics[width=\linewidth]{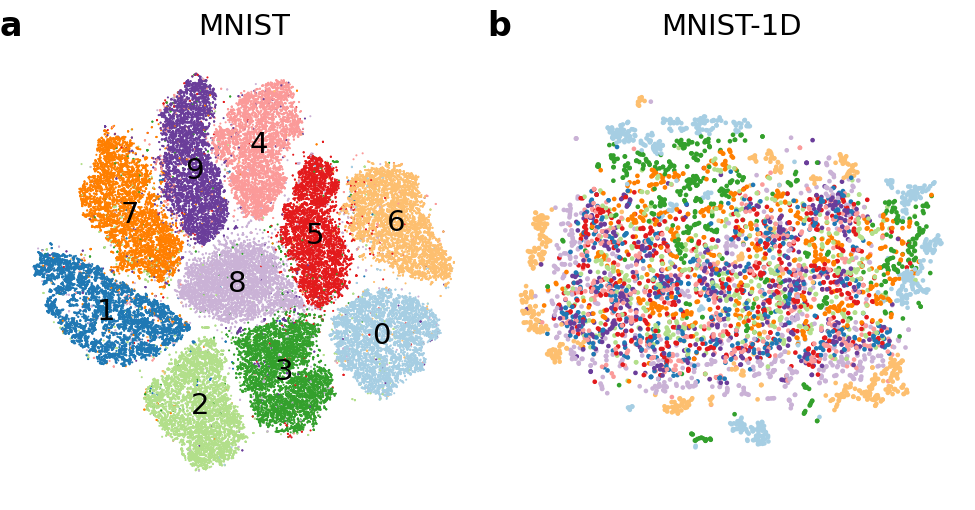}
    \caption{t-SNE visualization of MNIST and MNIST-1D datasets \cite{greydanus2020scaling}}
    \label{fig:dataset-cluster}
\end{figure}
Additionally, small datasets are often static and fixed in their size and structure, limiting researchers’ ability to explore variations in the data or introduce noise distributions. This lack of flexibility prevents models from being tested under varied conditions, making it harder to evaluate their robustness or adaptability in real-world scenarios where data distributions often change dynamically. Therefore, the uniform nature of small datasets like MNIST restricts the opportunity for models to showcase their generalization capabilities or handle more complex data.

In response to these limitations, there has been growing interest in exploring datasets that maintain the benefits of small size but offer more flexibility for experimentation. This allows for the study of model behavior under varying conditions, providing deeper insights into neural network performance, generalization, and efficiency. In particular, these types of datasets offer a valuable tool for researchers interested in rapid experimentation and understanding the underlying mechanisms of deep learning models. They help bridge the gap between small-scale prototyping and large-scale deployment by facilitating faster experimentation cycles and allowing researchers to study phenomena like overfitting, model complexity, and generalization without the need for extensive computational resources.

Ultimately, small-scale datasets play an essential role in advancing machine learning theory, particularly as they enable the investigation of deep learning questions in environments where scalability and rapid iteration are prioritized. By leveraging these datasets, researchers can conduct precise, controlled studies that inform larger-scale efforts, reducing the environmental impact and time required for experimentation. This approach helps strike a balance between the need for interpretable models and the push for high performance in real-world applications.

\section{Related Work}

Small-scale datasets have been pivotal in advancing deep learning research by providing accessible and computationally efficient benchmarks. Among these, the MNIST dataset\cite{lecun1998gradient} has become a gold standard for evaluating machine learning models, particularly in the context of image recognition. MNIST consists of 28x28 grayscale images of handwritten digits and has been extensively used to test the capabilities of various machine learning architectures, from simple linear models to complex deep neural networks. Its simplicity and small size make it ideal for fast experimentation and educational purposes, but this simplicity also limits its ability to distinguish between advanced architectures\cite{lecun1998gradient}.

Beyond MNIST, datasets like CIFAR-10 and CIFAR-100\cite{krizhevsky2009learning} have been employed to study model performance on more complex image data. Unlike MNIST, CIFAR datasets consist of 32x32 colored natural images, offering more challenging tasks for distinguishing between models. For example, studies have shown that CIFAR datasets are more effective at highlighting the differences in performance between Multi-Layer Perceptrons (MLPs) and Convolutional Neural Networks (CNNs), as well as among various CNN variants such as ResNets and vanilla CNNs\cite{he2016deep}. Despite their advantages, these datasets are still relatively small compared to modern large-scale datasets, and their structured nature limits their application in exploratory research or meta-learning scenarios, where smaller and more flexible datasets are preferable.

FashionMNIST\cite{xiao2017fashion} was introduced as an alternative to MNIST to address its limitations. FashionMNIST consists of 28x28 grayscale images of clothing items and provides a more challenging task for machine learning models. Unlike MNIST, where even simple models can achieve high accuracy, FashionMNIST presents more nuanced patterns, requiring advanced architectures to achieve competitive performance. However, while it introduces greater complexity, FashionMNIST remains similar in size and structure to MNIST, making it less suitable for certain research contexts where even smaller and more flexible datasets are required.

In the study conducted by \cite{bellman1959adaptive} noted that indeed smaller datasets are just 2D point clouds, devoid of spatial or temporal correlations between features and lacking manifold structures that a deep nonlinear classifier could use to escape the curse of dimensionality. 

For researchers focused on smaller datasets, Scikit-learn\cite{pedregosa2011scikit} provides several toy datasets, such as those for clustering or simple classification tasks. While useful for understanding foundational machine learning concepts, these datasets are not complex enough to evaluate the capabilities of deep learning architectures. They lack the sequential or hierarchical structures found in datasets like MNIST or CIFAR, limiting their applicability for studying advanced neural networks.

Greydanus et al.\cite{greydanus2020scaling} introduced MNIST-1D as a minimalist alternative to MNIST, addressing some of the limitations of existing small-scale datasets. MNIST-1D transforms the two-dimensional digit images of MNIST into one-dimensional time-series data, maintaining the essential features of the original dataset while significantly reducing its dimensionality. This design allows researchers to explore model performance under stricter computational constraints, making it particularly suitable for rapid experimentation and studies of inductive biases in neural networks. The dataset is procedurally generated, enabling researchers to introduce variations such as noise and distortions, which are often difficult to achieve with static datasets like MNIST or CIFAR.

In their work, Greydanus et al. demonstrated the utility of MNIST-1D by evaluating several standard models, including logistic regression, MLPs, CNNs, and GRUs. They observed that while CNNs performed well due to their ability to capture local spatial patterns, GRUs were slightly less effective due to MNIST-1D's emphasis on spatial rather than temporal relationships. Interestingly, MLPs achieved moderate success, highlighting the importance of non-linear architectures even in the absence of spatial priors. The authors also investigated the effects of data shuffling on model performance, showing that the removal of spatial structure significantly degraded accuracy across all models. These findings underscore the importance of inductive biases and model architecture in leveraging the structural features of MNIST-1D.

Additionally, Greydanus et al. explored the double descent phenomenon in neural networks, observing that increasing the size of the hidden layer in an MLP classifier led to an initial decline in performance, followed by a recovery as the model became over-parameterized. This behavior, consistent with findings in traditional CNNs, highlights the potential of MNIST-1D as a platform for studying broader questions in machine learning theory. They further examined the role of pooling methods in CNNs, showing that pooling without striding was highly effective in low-data regimes but less impactful when larger training sets were available.

Overall, MNIST-1D represents a significant step forward in addressing the limitations of traditional small-scale datasets. By combining the benefits of low dimensionality, flexibility, and procedural generation, it provides a versatile platform for studying advanced neural network architectures and their inductive biases. In this work, we aim to build on the foundation laid by Greydanus et al. by exploring additional architectures, such as ResNet, Temporal Convolutional Networks (TCNs), and Dilated Convolutional Neural Networks (DCNNs), to further understand the strengths and limitations of different models on this dataset.

\section{MNIST-1D Implementation}
\subsection{Dimensionality and Dataset Construction}

The MNIST-1D dataset, as introduced by Greydanus et al.\cite{greydanus2020scaling}, represents a novel approach to reimagining the MNIST dataset as one-dimensional time series data. This transformation was specifically designed to offer researchers a dataset that retains the complexity of traditional image datasets while requiring fewer computational resources for training. Unlike the original MNIST dataset, which consists of 28x28 grayscale images, MNIST-1D is characterized by 40-point 1D signals, offering new opportunities for exploring the impact of sequential data on machine learning models.

To construct MNIST-1D, the authors began by creating ten hand-crafted templates, each representing a digit from 0 to 9. These templates, as illustrated in Figure\ref{fig:black-signals}, were constructed using 12 predefined x-coordinates, providing a sparse but recognizable representation of the digits. To introduce variability and simulate real-world distortions, the dataset was augmented through several transformations. Each sequence was padded with between 36 and 60 zeros to extend the temporal structure and then circularly shifted by a random number of indices (up to 48). This circular shift allowed for variability in the temporal alignment of the signals.

To further enhance the realism of the dataset, Gaussian noise was added to each sequence. This noise introduced stochastic variations that mimicked distortions often present in practical applications. Gaussian smoothing was then applied, using a standard deviation of $\sigma = 2$, to introduce spatial correlations between neighboring points, making the signals more continuous and structured. Finally, the sequences were downsampled to 40 data points, resulting in the final digit-like signals shown in Figure\ref{fig:transformed-signals}.

This process ensured that MNIST-1D retained enough structure for meaningful classification tasks while introducing challenges related to noise, variability, and alignment. As depicted in Figure\ref{fig:white-signals}, these transformations allow MNIST-1D to serve as a robust platform for studying inductive biases in machine learning models, particularly those designed for sequential data.

\subsection{Benchmark Models and Results}

To establish benchmarks, Greydanus et al.\cite{greydanus2020scaling} implemented and evaluated four standard models: logistic regression, a fully connected multilayer perceptron (MLP), a 1D convolutional neural network (CNN), and a gated recurrent unit (GRU). Each model was trained using the Adam optimizer, with early stopping applied to select the best-performing model for evaluation. The test accuracy achieved by each model underscores the varying capabilities of these architectures in handling sequential data.

The logistic regression model achieved an accuracy of 32\%, demonstrating the limitations of linear models on MNIST-1D. This result highlights the need for more complex architectures to capture the spatial and sequential patterns inherent in the dataset. The MLP, benefiting from its non-linear activation functions and hidden layers, attained 68\% accuracy. However, its performance remained constrained by the absence of spatial or temporal priors.

 \begin{figure}[!htb]
   \centering
     \includegraphics[width=\linewidth]{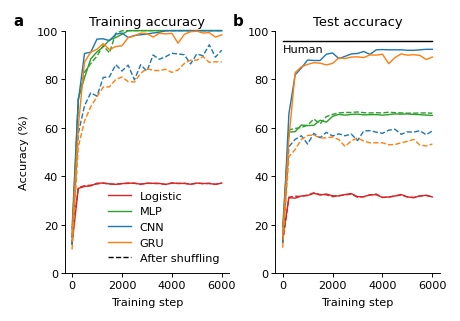}
   \caption{Train and test accuracy of various classification models on MNIST-1D \cite{greydanus2020scaling}}
   \label{fig:classification-accuracy}
\end{figure}
In contrast, the CNN achieved 94\% accuracy, showcasing the power of convolutional layers in capturing local spatial patterns. By applying convolutional filters, the CNN effectively learned features that generalized well across the dataset. The GRU, leveraging its ability to model temporal dependencies, reached 91\% accuracy. While slightly below the CNN, this performance indicates that temporal biases are also effective for MNIST-1D, though less critical than spatial biases.

Figure\ref{fig:classification-accuracy} presents the training and test accuracies of these models. The results clearly demonstrate the benefits of architectures that incorporate inductive biases tailored to the dataset's structure.

To analyze structural differences between MNIST and MNIST-1D, the original paper employed t-SNE for dimensionality reduction and visualization. The t-SNE projection of the MNIST dataset, shown in Figure \ref{fig:dataset-cluster}a, reveals ten distinct clusters, each corresponding to a digit class (0–9). This clustering indicates high-class separability in pixel space, allowing simple k-nearest neighbors (kNN) classifiers to achieve strong performance based on pixel-level similarities.

In contrast, the t-SNE visualization of MNIST-1D, presented in Figure \ref{fig:dataset-cluster}b (reproduced from Greydanus et al., 2024), shows a scattered distribution of data points without distinct clusters, implying that nearest neighbors in pixel space do not correspond to semantically similar classes. This lack of separability poses a challenge for traditional classifiers, as it requires models to learn complex representations beyond local pixel space proximity. Such properties make MNIST-1D similar to natural image datasets like CIFAR-10/100, where minimal pixel-space structure necessitates robust feature extraction. As discussed in the original paper, MNIST-1D’s design provides an effective benchmark for evaluating the robustness of feature extraction in models, especially those reliant on strong spatial or temporal priors.

\subsection{Human Benchmark}

In addition to evaluating machine learning models, the original paper included a human benchmark. A human expert trained on the MNIST-1D training set achieved an accuracy of 96\% on the test set. This performance provides a valuable upper bound for evaluating machine learning models, as it reflects the intuitive understanding of the dataset's structure by a human observer.

Interestingly, the per-digit error rates of the CNN closely aligned with those of the human expert, as shown in Figure\ref{fig:quickstart-accuracy}. This alignment suggests that the high accuracy achieved by the CNN was driven by meaningful statistical patterns in the data rather than overfitting to artifacts. Furthermore, the human benchmark underscores the interpretability and robustness of the MNIST-1D dataset, validating its utility as a research tool.

The MNIST-1D dataset offers a versatile and efficient platform for studying foundational machine learning concepts. Its unique combination of low dimensionality, variability, and structure enables researchers to evaluate the strengths and limitations of different model architectures under conditions that balance computational efficiency and experimental rigor.

\subsection{Metalearning and activation function}

The compact size of the MNIST-1D dataset enables researchers to explore more advanced metalearning optimization strategies that are computationally prohibitive on larger datasets. One notable example is the metalearning of activation functions, a novel area of research highlighted by the authors. Specifically, they parameterized their classifier’s activation function using a separate neural network, an MLP with the architecture 1→100→100→11→100→100→1 and tanhtanh activations. The output of this MLP was combined with an Exponential Linear Unit (ELU) function to introduce perturbations to the ELU’s shape, allowing the activation function to be dynamically adapted. The parameters of this MLP were optimized using meta-gradients through nested optimization, a technique that computes analytical gradients for both the primary model and the meta-learner.

The learned activation function significantly outperformed standard nonlinearities such as ReLU, ELU, and Swish, achieving a test accuracy improvement of over 5 percentage points. Interestingly, the resulting activation function exhibited a non-monotonic shape with two local extrema, a departure from traditional monotonic activation functions.

While prior research has explored activation function optimization (e.g.,\cite{clevert2015fast},\cite{ramachandran2017searching},  \cite{vercellino2017hyperactivations}), none have utilized analytical meta-gradients derived through nested optimization. Furthermore, earlier experiments, such as those by \cite{ramachandran2017searching}, required extensive multi-day training on large clusters of GPUs and TPUs. In contrast, this approach achieved comparable results within approximately one hour of CPU runtime, demonstrating a substantial reduction in computational resource requirements.

\subsection{Gradient-based metalearning}

The authors highlight that the primary objective of metalearning is to develop the ability to "learn how to learn." This is achieved through a two-level optimization framework: a fast inner loop, which operates using a conventional learning objective, and a slower outer loop, which updates meta-level properties of the learning process. A fundamental example of this approach is gradient-based hyperparameter optimization. Originally proposed by \cite{bengio2000gradient} and later extended to deep learning by\cite{maclaurin2015gradient}, this technique involves constructing a fully differentiable training loop, enabling backpropagation through the entire training process to optimize hyperparameters such as the learning rate and weight decay.

Despite its promise, scaling metalearning remains challenging. Metalearning algorithms are computationally intensive due to their nested optimization structure, often requiring substantial time and resources. Furthermore, these algorithms can become highly complex since most deep learning frameworks are not inherently designed for such recursive processes. Consequently, small-scale datasets, like MNIST-1D, are particularly valuable for developing and debugging metalearning algorithms, as they mitigate the computational burden while maintaining the complexity necessary for meaningful experimentation.

In this study, the authors implemented a gradient-based metalearning optimization for an MLP classifier on the MNIST-1D dataset, with the inner optimization loop explicitly defined using Stochastic Gradient Descent (SGD). The metalearning process successfully converged to an optimal learning rate of 0.62, regardless of the initial learning rate, whether excessively high or low. Notably, the entire optimization process was completed in approximately one minute on a CPU, underscoring the efficiency and practicality of this approach on small datasets.
\section{New Contributions}

To further demonstrate the utility of the MNIST-1D dataset in evaluating the strengths of various algorithms, we implemented models based on Residual Networks (ResNet), Temporal Convolutional Networks (TCN), and Dilated Convolutional Networks (DCNN) for classifying the images in the dataset. The details of our implementation are available in our GitHub repository: \url{https://github.com/michael-beebe/mnist1d/tree/master}\cite{mnist1d_github}.

\subsection{ResNet}

Residual Networks (ResNets)\cite{resnet} are a type of deep learning architecture that introduced the concept of residual learning to address challenges associated with training very deep neural networks. Traditionally, as networks become deeper, problems such as vanishing gradients and degradation of performance arise, making it difficult for the model to converge effectively. ResNets tackle these issues by incorporating skip connections, which allow the output of a layer to bypass one or more intermediate layers and be added directly to the output of subsequent layers. This enables the network to learn residual mappings rather than trying to directly map the input to the output, significantly improving training stability and performance.

What makes ResNet unique is the use of these skip connections, which essentially reformulate the learning problem. Instead of forcing layers to learn the full transformation, the network learns the difference (or residual) between the input and the desired output of the block. This simple yet powerful idea allows very deep networks—sometimes with hundreds or even thousands of layers—to be trained effectively. Additionally, ResNets maintain the same dimensionality between layers within a block, simplifying the addition of skip connections and preserving feature alignment.

The primary benefits of ResNet include improved gradient flow during backpropagation, reduced risk of overfitting for very deep networks, and enhanced convergence speed. The architecture’s ability to train deeper networks allows it to capture more complex hierarchical features, making it particularly well-suited for tasks such as image recognition, where features at multiple scales are critical. Furthermore, its modular design, consisting of residual blocks, enables flexibility and adaptability across a variety of data types and applications.

For the MNIST-1D dataset, ResNet is a natural choice due to its ability to effectively model hierarchical features in the data. While the MNIST-1D dataset is simpler than traditional image datasets, it still exhibits sequential patterns and dependencies that benefit from the hierarchical feature extraction capabilities of ResNet. The skip connections in ResNet also provide robustness against potential overfitting, which is particularly important when dealing with smaller datasets like MNIST-1D.

We hypothesize that the ResNet architecture will perform well on this dataset, as it can leverage its deep hierarchical structure to effectively capture sequential dependencies in the data. Additionally, the skip connections should allow the model to generalize better compared to standard convolutional architectures, particularly when training with a limited number of samples. We expect ResNet to outperform simpler models by effectively extracting and combining both low-level and high-level features from the dataset.

\subsection{ResNet Implementation}

Our implementation of ResNet is a one-dimensional adaptation of the ResNet architecture designed specifically for sequential data. This architecture follows the principles of residual learning by incorporating skip connections to enable effective training of deeper networks. Below, we explain the key components of our implementation.

The model begins with an initial convolutional layer that takes a one-dimensional input sequence with a single channel and outputs a specified number of hidden channels. This convolutional layer uses a kernel size of 7, a stride of 2, and padding of 3. This configuration ensures that the input sequence is downsampled by half while capturing a wide context of features. The output is then passed through a batch normalization layer, which normalizes the activations to stabilize training, followed by a ReLU activation function to introduce non-linearity.

The core of the model is a residual block, consisting of two convolutional layers, each with a kernel size of 3, a stride of 1, and padding of 1. Both convolutional layers maintain the same spatial dimensions, ensuring that the input and output sequences remain aligned. Batch normalization is applied after each convolution to enhance stability during training. The output of the first layer in the residual block is stored as the "identity," which is then added to the output of the second convolutional layer, forming the skip connection. This skip connection enables the model to learn residual mappings, which mitigate the vanishing gradient problem and improve convergence in deeper networks. The combined output is further passed through a ReLU activation function.

Following the residual block, the model applies global average pooling, which reduces the spatial dimensions of the feature map to a fixed size of 1. This operation aggregates information across the entire sequence, resulting in a feature vector of size equal to the number of hidden channels. The global average pooling layer ensures that the model is compatible with variable-length input sequences while focusing on the most salient features.

Finally, the pooled feature vector is passed through a fully connected linear layer that maps the hidden representation to the desired output size. This layer enables the model to perform tasks such as classification or regression, depending on the application. The output of the model represents the predicted values for the given task.

This implementation of ResNet is well-suited for processing sequential data such as MNIST-1D. The use of residual connections allows the model to effectively capture hierarchical features while maintaining robustness against overfitting. The global average pooling layer ensures that the model remains efficient and adaptable, even for datasets with varying input sequence lengths. Overall, our ResNet implementation combines the strengths of residual learning and 1D convolutional processing to achieve high performance on sequential datasets.

\subsection{Temporal Convolutional Neural Network (TCN)}

Temporal Convolutional Networks (TCNs)\cite{tcn} are a class of convolutional neural networks specifically designed for processing sequential data. Unlike traditional recurrent architectures such as LSTMs or GRUs, TCNs rely purely on convolutional operations to model temporal dependencies, making them faster to train and easier to parallelize. TCNs leverage causal convolutions, which ensure that the output at any time step depends only on the current and past inputs, preserving the temporal order of the sequence.

What makes TCNs unique is their use of dilated convolutions, which allow the network to capture long-range dependencies without increasing the number of layers. In dilated convolutions, the filter skips input values at regular intervals determined by the dilation rate. This mechanism enables TCNs to achieve a larger receptive field while maintaining computational efficiency. Additionally, TCNs employ residual connections, which help mitigate the vanishing gradient problem and allow for deeper network architectures by ensuring stable gradient flow during backpropagation.

The benefits of TCNs include their ability to model both local and global dependencies in sequential data, their robustness to vanishing gradients, and their efficiency in training compared to recurrent networks. Because TCNs rely solely on convolutions, they are inherently parallelizable and can process sequences much faster than RNNs, especially on large datasets or hardware optimized for convolutions. Furthermore, the use of residual blocks improves the network’s ability to generalize by allowing it to learn the difference (residual) between the input and output of a block.

TCNs are well-suited for the MNIST-1D dataset due to their ability to capture sequential dependencies and extract hierarchical patterns from the data. The dataset consists of 1D sequences, where understanding the relationships between adjacent and distant elements is critical for accurate classification. The use of dilated convolutions allows TCNs to capture patterns over varying time scales, making them particularly effective for this type of data.

We hypothesize that TCNs will perform well on the MNIST-1D dataset, leveraging their hierarchical feature extraction and long-range dependency modeling capabilities. Additionally, the residual connections in TCNs are expected to improve the network’s robustness and ability to generalize, particularly in scenarios with limited training data.

\subsection{TCN Implementation}

Our implementation of the Temporal Convolutional Network (TCN) is designed to process sequential data by leveraging residual connections and hierarchical feature extraction, making it well-suited for the MNIST-1D dataset. Below, we describe the key components of our TCN implementation.

The model begins with an initial convolutional layer that takes a one-dimensional input sequence with a single channel and outputs the specified number of channels for the first residual block. This layer uses a kernel size of 7, a stride of 2, and padding of 3, which allows it to downsample the input sequence while capturing a broad range of features. The output of this layer is normalized using batch normalization and passed through a ReLU activation function, ensuring stability and non-linearity early in the network.

The core of the model is composed of multiple residual blocks, organized in a series of stacked layers. Each block consists of two convolutional layers with a kernel size of 3, a stride of 1, and padding of 1, ensuring the input and output sequences remain aligned in size. These layers are followed by batch normalization and ReLU activation, promoting efficient training. A dropout layer is added after each convolutional operation to regularize the network and prevent overfitting, especially when training on smaller datasets like MNIST-1D. 

A key feature of the residual blocks is the use of skip connections. At each block, the input (referred to as the "identity") is added to the output of the second convolutional layer. If the dimensions of the input and output differ, a 1D convolutional layer is applied to the identity to match the channel dimensions. This addition ensures that the network learns residual mappings, enabling deeper architectures without the risk of vanishing gradients.

After processing the input through all residual blocks, the model applies global average pooling. This reduces the dimensionality of the feature maps to a single value per channel by computing the mean across the sequence length. The resulting fixed-size feature vector is then passed to a fully connected linear layer, which maps it to the desired output size for classification tasks.

Our TCN implementation is well-suited for the MNIST-1D dataset due to its ability to handle sequential data and capture long-range dependencies. The use of residual connections ensures efficient training and robust performance, while the hierarchical feature extraction facilitates learning both local and global patterns in the data. By incorporating dropout and batch normalization, the model achieves a balance between capacity and generalization, making it effective for tasks involving smaller datasets.

\subsection{Dilated Convolutional Neural Network (DCNN)}

Dilated Convolutional Neural Networks (DCNNs)
\cite{dcnn} are a variant of traditional convolutional networks that use dilated (or atrous) convolutions to expand the receptive field without increasing the number of parameters or reducing the spatial resolution of the input. In a dilated convolution, the filter skips input values at fixed intervals (determined by the dilation rate), enabling the network to efficiently model long-range dependencies in data while maintaining computational efficiency.

What makes DCNNs unique is their ability to aggregate multiscale information by increasing the receptive field exponentially with respect to the depth of the network. This allows DCNNs to capture both fine-grained local details and broader global patterns simultaneously, making them particularly effective for tasks where such multiscale features are important. Unlike traditional convolutional layers, dilated convolutions achieve this without downsampling or adding additional pooling layers, preserving the input's resolution throughout the network.

The primary benefits of DCNNs include their efficiency in capturing long-range dependencies, their ability to handle hierarchical features, and their robustness in modeling sparse or structured data. By combining local and global feature extraction in a single framework, DCNNs are capable of understanding complex patterns in sequential or spatial data. Moreover, their computational efficiency makes them suitable for applications where memory or processing power is limited.

DCNNs are well-suited for the MNIST-1D dataset because the data consists of sequential patterns where both local and global relationships between features are critical for accurate classification. The dilated convolutional layers in DCNNs enable the model to recognize these relationships effectively across different scales, allowing the network to generalize well even with relatively simple datasets like MNIST-1D.

We hypothesize that DCNNs will achieve strong performance on the MNIST-1D dataset due to their multiscale feature extraction capabilities and efficient handling of long-range dependencies. Additionally, the preservation of input resolution throughout the network is expected to improve classification accuracy by retaining critical details in the data that are essential for distinguishing between classes.

\subsection{DCNN Implementation}

Our implementation of the Dilated Convolutional Neural Network (DCNN) is specifically designed to leverage dilated convolutions for hierarchical feature extraction and efficient handling of long-range dependencies in sequential data. Below, we describe the architecture and functionality of our DCNN implementation.

The model begins with three stacked dilated convolutional layers. Each layer applies a 1D convolution operation with a kernel size of 3, a dilation rate of 2, and padding equal to the dilation rate. This configuration ensures that the receptive field of the network expands exponentially with each layer, enabling the model to capture long-range dependencies in the data without increasing the computational complexity. The first convolutional layer maps the input sequence, which has a single channel, to a higher-dimensional space defined by the number of channels. Subsequent layers maintain this channel dimensionality while progressively aggregating multiscale features.

Each convolutional layer is followed by a ReLU activation function to introduce non-linearity, ensuring that the network can model complex relationships in the data. The use of stacked dilated convolutions allows the model to effectively combine local and global patterns, a key requirement for tasks involving sequential datasets such as MNIST-1D.

To reduce the dimensionality of the output from the convolutional layers and create a fixed-size feature vector, the model employs a global average pooling layer. This layer computes the average value across the sequence length for each channel, reducing the output to a vector of size equal to the number of channels. This operation ensures that the network can handle variable-length inputs while focusing on the most salient features in the sequence.

Finally, the feature vector is passed through a fully connected linear layer, which maps the output to the desired number of classes for classification tasks. This architecture enables the model to learn from both fine-grained local details and broader global patterns in the data, making it highly effective for tasks requiring sequential feature extraction.

The DCNN is particularly well-suited for the MNIST-1D dataset due to its ability to capture hierarchical and long-range dependencies. The use of dilated convolutions ensures that the model can efficiently extract multiscale features, which are critical for distinguishing between different classes in the dataset. Additionally, the simplicity of the architecture, combined with its computational efficiency, makes it an excellent choice for smaller datasets like MNIST-1D.

\section{Experimental Results}

This section analyzes the test accuracy of various models trained on the MNIST-1D dataset over different training steps: 1,000, 2,000, and 10,000. The models used in the experiments include Logistic Regression, Multi-Layer Perceptron (MLP), Convolutional Neural Networks (CNN), Recurrent Neural Networks (GRU), Residual Networks (ResNet), Temporal Convolutional Networks (TCN), and Dilated CNNs. Human performance is used as a benchmark to contextualize the performance of the models.

\begin{figure}[!htb]
    \centering
    \includegraphics[width=\linewidth]{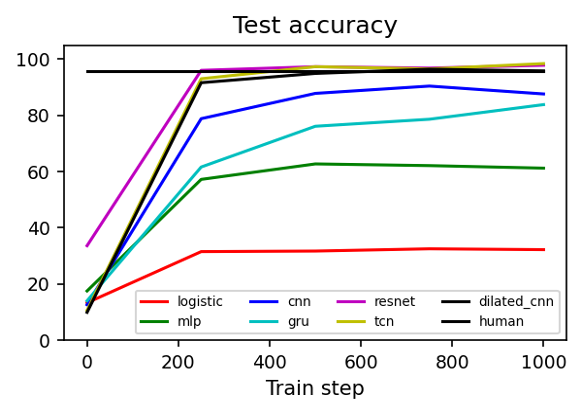}
    \caption{Test accuracy over training steps with new models\\\centering(1,000 steps)}
    \label{fig:results-1000-steps}
\end{figure}

Figure\ref{fig:results-1000-steps} illustrates the test accuracies of the models after 1,000 training steps. The Logistic Regression model achieves the lowest accuracy, as expected, due to its linear nature and inability to model the non-linear relationships inherent in the MNIST-1D dataset. The Multi-Layer Perceptron (MLP) performs better, leveraging its hidden layers and non-linear activation functions to capture more complex patterns.

The CNN demonstrates a significant improvement over the MLP, as its convolutional layers are adept at extracting local spatial patterns from the data. Similarly, the GRU, a recurrent neural network, performs well by modeling temporal dependencies in the sequential data. However, both models are outperformed by ResNet, which utilizes residual connections to enable deeper architectures and better gradient flow.

The TCN and Dilated CNN achieve the highest accuracies at this stage, showcasing their ability to effectively capture long-range dependencies and hierarchical patterns in the data. The TCN leverages dilated convolutions and residual connections to process sequential data efficiently, while the Dilated CNN expands its receptive field without adding computational overhead. Both models approach or exceed human-level accuracy even within 1,000 training steps.

\begin{figure}[!htb]
    \centering
    \includegraphics[width=\linewidth]{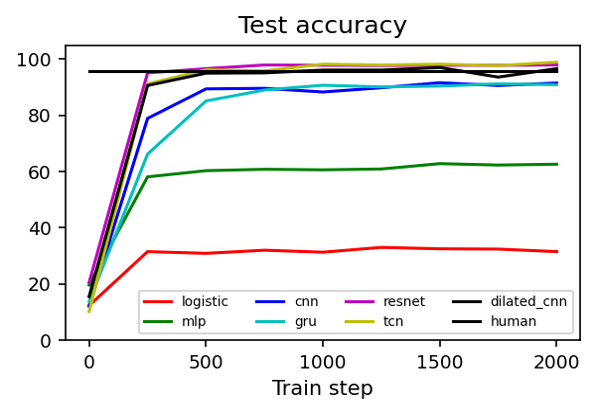}
    \caption{Test accuracy over training steps with new models\\\centering(2,000 steps)}
    \label{fig:results-2000-steps}
\end{figure}

As shown in Figure~\ref{fig:results-2000-steps}, increasing the training steps to 2,000 results in noticeable accuracy improvements across all models. The Logistic Regression model sees minimal gains, reinforcing its limitations in capturing complex data relationships. The MLP achieves modest improvements, but its simpler architecture prevents it from competing with more advanced models.

The CNN and GRU models benefit more substantially from the additional training, as their architectures are better suited for capturing patterns in the MNIST-1D dataset. ResNet continues to outperform both, showcasing its advantage in handling deeper networks with residual connections. The gap between ResNet and the simpler models widens as the training steps increase.

The TCN and Dilated CNN maintain their superior performance, with both models nearing perfect accuracy. Their ability to model multiscale features and long-range dependencies enables them to generalize effectively, even with relatively small training datasets like MNIST-1D.

\begin{figure}[!h]
    \centering
    \includegraphics[width=\linewidth]{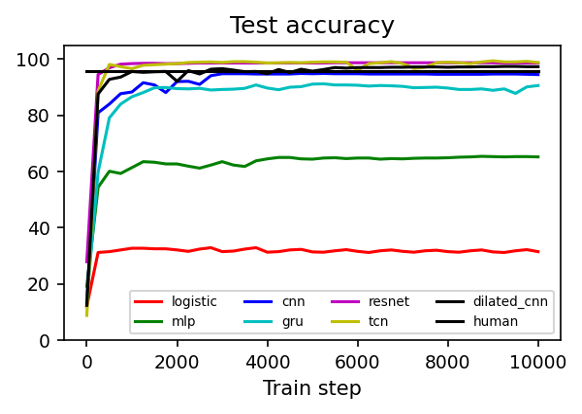}
    \caption{Test accuracy over training steps with new models\\\centering(10,000 steps)}
    \label{fig:results-10000-steps}
\end{figure}

At 10,000 training steps, as shown in Figure\ref{fig:results-10000-steps}, most models reach their performance ceilings. Logistic Regression and MLP show minimal improvement beyond 2,000 steps, indicating that their capacity to model the dataset’s complexity is limited. The CNN and GRU models plateau as well, although at a higher accuracy than the simpler models.

ResNet continues to improve, benefiting from its depth and residual learning capabilities. However, the TCN and Dilated CNN achieve the highest accuracies overall, consistently outperforming the other models. Their ability to capture both local and global dependencies across different training steps demonstrates their robustness and adaptability. Notably, the top-performing models meet or exceed human-level performance, validating the efficacy of their architectures for this dataset.

\subsection{Summary of Results}

This section underscores several important observations about the performance of different classification models on the MNIST-1D dataset. First, model complexity plays a crucial role in achieving high accuracy. Advanced architectures, such as Temporal Convolutional Networks (TCN), Dilated CNNs, and Residual Networks (ResNet), consistently outperformed simpler models like Logistic Regression and Multi-Layer Perceptrons (MLPs). This highlights the necessity of selecting architectures that align with the dataset's sequential and hierarchical structure to fully exploit its features.

Another key finding is the significant impact of inductive biases on model performance. Models that incorporate biases suitable for sequential data, such as convolutional layers in CNNs, ResNet, and Dilated CNNs, or temporal dependencies in GRU and TCN, demonstrated substantially higher accuracies. These biases enable the models to effectively capture the underlying structure of the dataset, providing a distinct advantage over models without such specialized mechanisms.

The results also demonstrate the benefits of extended training. While all models showed improvements with additional training steps, the rate of accuracy gains diminished as simpler models approached their capacity limits. In contrast, advanced architectures like ResNet, TCN, and Dilated CNN continued to improve, achieving the highest accuracies even after extended training. Notably, these models consistently exceeded human benchmarks, showcasing their capability to generalize effectively on the MNIST-1D dataset.

Among the models tested, TCN and Dilated CNN stood out for their superior performance. Their ability to model long-range dependencies and multiscale features allowed them to consistently achieve the highest accuracies across all training steps. These results validate the hypothesis that architectures tailored to the dataset's characteristics are particularly effective and emphasize the importance of leveraging advanced inductive biases to achieve state-of-the-art performance.

Overall, these findings validate the efficacy of using MNIST-1D as a benchmark for studying the interplay between model complexity, inductive biases, and training dynamics. They also reinforce the importance of architectural innovations in improving performance on small, structured datasets.

\section{Conclusion}

This work builds on the foundational study of the MNIST-1D dataset, highlighting its potential as a robust platform for exploring the strengths and limitations of diverse machine learning architectures. We began by reviewing the construction and benefits of MNIST-1D, noting its transformation from the traditional MNIST dataset into one-dimensional time-series data that retain critical features while enabling efficient experimentation. This dataset provides researchers with a unique opportunity to investigate model performance under computational constraints and explore inductive biases in neural networks.

Our evaluation included several advanced architectures—ResNet, Temporal Convolutional Networks (TCN), and Dilated Convolutional Neural Networks (DCNN)—in addition to previously benchmarked models like logistic regression, MLPs, CNNs, and GRUs. Each of these architectures leverages distinct mechanisms to capture sequential and spatial patterns, with ResNet focusing on residual learning, TCN emphasizing long-range dependencies through dilated convolutions, and DCNN achieving multiscale feature aggregation. These models were implemented and tested on MNIST-1D, yielding results that align with our hypotheses regarding their suitability and effectiveness for this dataset.

Experimental findings revealed the superior performance of advanced architectures like TCN and DCNN, which consistently outperformed simpler models such as logistic regression and MLP across various training steps. These architectures demonstrated their ability to capture both local and global dependencies, validating the importance of incorporating inductive biases tailored to the dataset's structure. ResNet, with its robust residual learning framework, also achieved strong results, emphasizing the value of hierarchical feature extraction for sequential data.

Through this study, we observed that models with advanced architectural designs benefit most from the MNIST-1D dataset's characteristics, such as its sequential structure and noise-induced variability. These insights contribute to our understanding of how different neural network architectures perform on datasets with reduced dimensionality and resource demands. Additionally, the use of MNIST-1D as a benchmark reinforces its utility in bridging the gap between small-scale research and real-world applications.

Future work will extend this analysis by exploring additional architectures, optimizing hyperparameters specific to MNIST-1D, and introducing variants of the dataset to further challenge the generalization capabilities of these models. By continuing to investigate the relationship between inductive biases and performance on structured datasets, we aim to advance the development of deep learning models optimized for environments with computational constraints.

\bibliographystyle{IEEEtran}
\bibliography{main}


\end{document}